\pdfoutput=1
\documentclass[11pt]{article}

\usepackage{emnlp2021}

\usepackage{times}
\usepackage{latexsym}

\usepackage[T1]{fontenc}
\usepackage[utf8]{inputenc}

\usepackage{microtype}
\usepackage{graphicx} 
\usepackage{subfig}
\usepackage{amsmath}

\usepackage{algorithm} 
\usepackage{algpseudocode}

\usepackage{xcolor}
\algnewcommand\True{\texttt{True}\space}
\algnewcommand\False{\texttt{False}\space}

\usepackage{xspace}
\makeatletter
\DeclareRobustCommand\onedot{\futurelet\@let@token\@onedot}
\def\@onedot{\ifx\@let@token.\else.\null\fi\xspace}
 
\def\ie{\emph{i.e}\onedot}

\makeatother

\RequirePackage{color}

\definecolor{dact-color}{RGB}{55,88,218}
\definecolor{deebert-color}{RGB}{255,180,0}
\definecolor{pabee-color}{RGB}{0,128,0}

\newcommand*{\affaddr}[1]{#1} % No op here. Customize it for different styles.
\newcommand*{\affmark}[1][*]{\textsuperscript{#1}}
\newcommand*{\email}[1]{\texttt{#1}}
\renewcommand\footnotemark{}
\title{DACT-BERT: Differentiable Adaptive Computation Time for an Efficient BERT Inference}

\author{%
Cristóbal Eyzaguirre\affmark[1,*]
, Felipe del Río\affmark[1], Vladimir Araujo\affmark[1,2], Alvaro Soto\affmark[1]\\
\affaddr{\affmark[1]Pontificia Universidad Católica de Chile},
\affaddr{\affmark[2]KU Leuven}\\
\email{ceyzagui@stanford.edu},
\email{\{fidelrio, vgaraujo\}@uc.cl}, 
\email{asoto@ing.puc.cl}
\thanks{\affmark[*]Work done while at Pontificia Universidad Católica de Chile.}
}

\begin{document}
\maketitle
\begin{abstract}
Large-scale pre-trained language models have shown remarkable results in diverse NLP applications. Unfortunately, these performance gains have been accompanied by a significant increase in computation time and model size, stressing the need to develop new or complementary strategies to increase the efficiency of these models.
% current large language models, such as BERT. 
In this paper we propose DACT-BERT, a differentiable adaptive computation time strategy for BERT-like models.
DACT-BERT adds an adaptive computational mechanism to BERT's regular processing pipeline, which controls the number of Transformer blocks that need to be executed at inference time.
% DACT-BERT adds an adaptive computation mechanism to the regular processing pipeline of BERT, a mechanism that controls the number of Transformer blocks that BERT needs to execute at inference time. 
By doing this, the model learns to combine the most appropriate intermediate representations for the task at hand.
% With respect to previous works, our method has the advantage of being fully differentiable and directly integrated to BERT's main processing pipeline. 
% This enables the incorporation of gradient-based transparency mechanisms to improve interpretability. 
% Furthermore, by discarding useless steps, DACT-BERT facilitates the understanding of the underlying process used by BERT to reach an inference. 
% Our experiments demonstrate that our approach performance is competitive in comparison to the baselines but excels on reduced computational regimes.\fdr{arreglar esta ultima frase}
Our experiments demonstrate that our approach, when compared to the baselines, excels on a reduced computational regime and is competitive in other less restrictive ones.
% when significantly reducing computation.
% , and is competitive for small reductions.
% Additionally, they also demonstrate that DACT-BERT helps to improve model interpretability.
\end{abstract}

\section{Introduction}

The use of pre-trained language models based on large-scale Transformers \citep{vaswani2017attention} has gained popularity after the release of BERT \citep{devlin-etal-2019-bert}.
% mainly due to its success to support a large variety of NLP tasks~\citep{BERTology:2020}. 
% In particular, BERT \citep{devlin-etal-2019-bert} has become one of the most popular tools. 
The usual pipeline consists of finetuning BERT by adapting and retraining its classification head to meet the requirements of a specific NLP task. Unfortunately, the benefits of using a powerful model are also accompanied by a highly demanding computational load. 
In effect, current pre-trained language models such as BERT have millions of parameters, making them computationally intensive both during training and inference. 
% Furthermore, the large number of parameters makes these models hard to analyze, limiting their interpretability.  

While high accuracy is usually the ultimate goal, computational efficiency is also desirable.
% and model interpretability are also desirable objectives.
% In terms of computational efficiency, t
The use of a demanding model not only causes longer processing times and limits applicability to low-end devices, but it also has major implications in terms of the environmental impact of AI technologies~\citep{GreenAI:2019}. 
As an example, \citet{EnergyCostIA:McCallum:2019} provides an estimation of the carbon footprint of 
several large NLP models, including BERT, concluding that they are becoming unfriendly to the environment. 
% In terms of interpretability, potential biased or malicious uses of AI technologies, in particular NLP applications, are increasing the need to provide them with the ability to explain their decisions~\citep{XAI:2018}. As an example, \citep{EU:Regulations:2016} analyzes the implications of a new legislation of the European Union that enforces the right to explanation on algorithmic decision-making. 

Fortunately, recent works have shown that behind BERT's immense capacity, there is considerable redundancy and over-parametrization \citep{kovaleva-etal-2019-revealing,BERTology:2020}. 
Consequently, others works have explored strategies to develop efficient and compact versions of BERT. 
% A relevant one consists of distilling the knowledge from a pre-trained model into a smaller network \citep{sanh2020distilbert,jiao2020tinybert}. The main disavantage of this approach is the inherent 
% complexity of conducting an efficient and effective distillation process. 
One such strategy known as dynamic Transformers consists of providing BERT with an adaptive mechanism to control how many Transformers blocks are used \citep{xin2020deebert,liu2020fastbert,zhou2020bertlosespatience}. 
% A common limitation of these adaptive models is that they all rely heavily on an external hyperparameter that must be carefully tuned for each model and task.

In this paper, we present DACT-BERT, an alternative to current dynamic Transformers that uses an Adaptive Computation Time mechanism~\citep{Graves2016AdaptiveCT} to control the complexity of the processing pipeline of BERT. 
This mechanism controls the number of Transformer blocks executed at inference time by using additional classifiers. 
This allows resulting models to take advantage of the information already encoded in intermediate layers without the need to run all layers.
Specifically, our model integrates DACT, a fully  differentiable variant of the adaptive computation module ~\citep{eyzaguirre2020differentiable} that allows us to train a halting neuron after each Transformer block. This neuron indicates the confidence the model has on returning the correct answer after executing said block.
We use the DACT algorithm to determine when the answer stabilizes in a given output using the halting neuron, and halt once we are sure running more blocks cannot change the output.
% Based on this, the computation can be stopped when the response predicted by the model stabilizes in a given output, making unnecessary to continue running the processing pipeline. 

% With respect to previous works, our method has the advantage of being fully differentiable and 
% directly integrated into BERT's main processing pipeline. This avoids complexities associated to 
% the calibration of external hyperparameter that are task dependent. 
% In terms of interpretability, by discarding useless steps, DACT-BERT facilitates the understanding of the underlying process used by BERT to reach each inference.
% Furthermore, this enables the incorporation of gradient-based 
% transparency mechanisms~\citep{integratedgradients} to improve interpretability. Our experiments 
% using tasks from the GLUE benchmark~\citep{wang-etal-2018-glue} demonstrate that our approach is 
% effective in significantly reducing computational complexity without affecting model accuracy. 
% Additionally, they also demonstrate that DACT-BERT helps to improve model interpretability by 
% explaining the relationship between not only predictions and inputs, but also the model confidence score and 
% inputs. 

\vspace{-0.1cm}
\section{Related Work}
\vspace{-0.1cm}
% \section{Efficient Transformers}

% Recently research has attempted to improve the efficiency of Transformer-based language models like BERT. According to \citep{zhou2020bertlosespatience}, approaches can be categorized in two groups. 
% The first group, \textit{static approaches}, contains models designed to be more compact and are trained by extracting knowledge from pre-trained models.
% The second approach instead reuses the previously trained model, but allows it to dynamically choose different computational paths for each input instance during inference. 
% Our method DACT-BERT falls into the latter category.
Several architectures have been designed to avoid overcomputing in Transformer-based models.
% According to \citet{zhou2020bertlosespatience}, approaches can be categorized in two groups. 
According to \citet{zhou2020bertlosespatience}, there are two groups.

\subsection{Static Efficient Transformers}
\vspace{-0.1cm}

% One such strategy is to use lightweight architectures that are trained from scratch. 
% As an example, 
% ALBERT \citep{Lan2020ALBERT} proposes cross-layer parameter sharing as a way to improve model efficiency.
% Similar methodologies have also been previously explored by \citet{NIPS2019_8358,dehghani2018universal}, demonstrating the effectiveness of weight-tied Transformers.

% A second strategy is to distill the knowledge of pretrained models into a more compact.
% \fdr{Comenté la parte de ALbert mas por que creo que es una estrategia para reducir memoria ás que computo}
The first strategy is to distill the knowledge of pretrained models into more efficient "students".
Models such as PKD-BERT \citep{sun-etal-2019-patient}, TinyBERT \citep{jiao2020tinybert}, and DistilBERT \citep{sanh2020distilbert} compress the knowledge of large models (teachers) into more compact or efficient ones to obtain similar performance at a reduced computation or memory cost.
While these approaches effectively reduce the total calculation needed to execute the model,
% (measured in FLOPs or number of parameters)
they are limited in the same way as BERT, they do not take into account that some examples could be less complicated than others and always use the same amount of computation. 
% Furthermore, the application of a suitable distillation process is not a trivial task that introduces extra complications.

\subsection{Dynamic Transformers}
\vspace{-0.1cm}

Recently, a series of algorithms have been proposed to reduce computation in Transformer language models based on early exiting \citep{kaya2019shallow}. 
Models such as DeeBERT \citep{xin2020deebert}, FastBert \citep{liu2020fastbert}, and PABEE \citep{zhou2020bertlosespatience} introduce intermediate classifiers after each Transformer block. These classifiers are then trained independently from these blocks (not end-to-end).
At inference, a ``halting criterion'' is used to dynamically determine the number of blocks needed to perform a specific prediction. Instead of using a brittle confidence approach \citep{guo2017calibration} to determine when to stop, recent approaches rely on computing the Shannon's entropy of the output probabilities \citep{xin2020deebert,liu2020fastbert}, an agreement between intermediate classiﬁers \citep{zhou2020bertlosespatience}, or a trained confidence predictor \citep{xin-etal-2021-berxit}.
% or counting the number of times intermediate classiﬁers have agreed on an output \citep{zhou2020bertlosespatience}.
% A limitation with these approaches is that they are non-differentiable model and use a fix metric which is not adapted for the purpose. 
% \textbf{Agregar: BERxiT es mas reciente y es equivalente a DeeBERT}

% In contrast to previous works, we propose a fully differentiable alternative to achieve adaptive computation in Transformers based architectures. Our method takes inspiration from DACT \citep{eyzaguirre2020differentiable}, a technique proposed for visual reasoning tasks. 
% Unlike previous works that use heuristic proxies of models confidence to decide when to halt, DACT-BERT learns this behavior.
% In addition, we can train the complete model, adapting the weights of the Transformer blocks in the process, because it is totally differentiable.
Unlike previous works that use heuristic proxies of models confidence to decide when to halt, DACT-BERT is based on a learning scheme that induces the model to halt when it predicts that its current answer will not change with further processing.  As an illustrative example consider a difficult input. Here, our model could “understand” that further processing steps are superfluous and decide to stop early, even if its current answer has a low confidence. On the other hand, existing early stopping models would keep wasting computation because the confidence is low.

\vspace{-0.1cm}
\section{DACT-BERT: Differentiable Adaptive Computation Time for BERT}
\vspace{-0.1cm}

Dynamic early stopping methods use a proxy of model confidence to decide when it is safe to cut computation. In this work our signaling module, DACT, approximates this gating mechanism with a soft variant that allows our model to independently learn the confidence function.
% produces a prediction based on intermediate results and the confidence it has on them.
This mechanism can then be used to detect when stable results are obtained, allowing for the reduction of the total number of steps necessary for a given prediction. The original formulation of DACT \citep{eyzaguirre2020differentiable} applies this module to recurrent models. In our case, we adapt the formulation to the case of Transformer based architectures, namely BERT. 

\begin{figure*}[h]
\begin{center}
\includegraphics[width=0.8\textwidth]{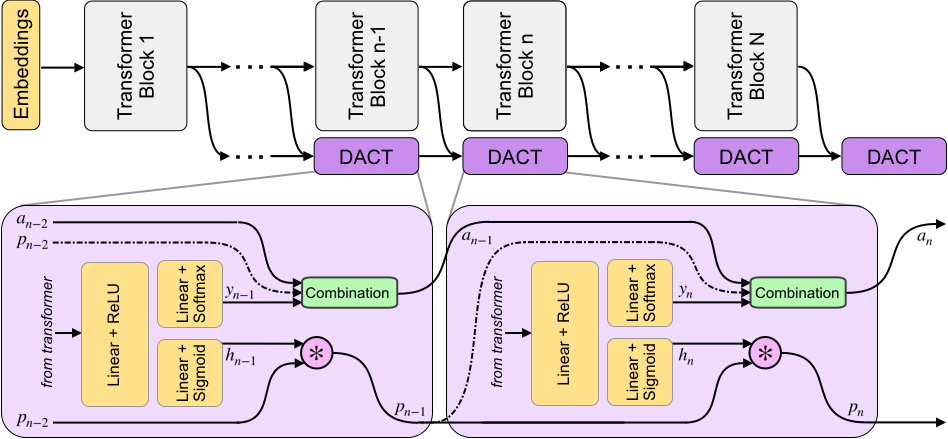}
\end{center}
% \vspace{-0.25cm}
\caption{DACT-BERT adds an additional classification layer after each Transformer block, along with a sigmoidal \textit{confidence function}. DACT-BERT combines the Transformer hidden state and the outputs and confidences of all earlier layers into an accumulated answer $a_n$. Later, during inference, the model is halted once $a_n \approx a_{N}$.}
\label{diagram}
\vspace{-0.25cm}   
\end{figure*}

% The original DACT paper applied this module to recurrent models, but it's formulation makes it straightforward to adapt it into sequential deep models such as Transformers. 

As shown in Figure~\ref{diagram}, DACT-BERT introduces additional linear layers after each computational unit, similar to the \textit{off-ramps} in DeeBERT \citep{xin2020deebert} or the student classifiers in the work of \citet{liu2020fastbert}.
% As in both these cases, we define the discrete unit of computation to be a single BERT Transformer block.
However, differently from previous approaches, each $n$-th DACT module also computes an scalar confidence score (or halting value) $h_n$ in addition to the output vector $y_n$. Following \citet{devlin-etal-2019-bert}, both, $y_n$ and $h_n$, are estimated by using the classification token ($[CLS]$) that is included in BERT as part of the output representation of each layer.
During training all the output vectors and halting values are combined to obtain $a_n$ and the final predicted probabilities $p_n$ following an expression that can be rewritten as the weighted average of all intermediate outputs $y_n$ multiplied by a function of the confidences of earlier blocks.
Then, during inference, the confidence scores can be used to reduce computation.

% Auxiliary accumulator variables $a_n$ are used to build the full output of the model, $a_N$, by accumulating intermediate outputs $y_n$:
% \begin{equation}
%     \label{eq:2}
%     a_n = \begin{cases}
%     \overrightarrow{0} \qquad \text{if} \quad n=0\\
%      y_n p_{n-1} + a_{n-1} \left( 1 - p_{n-1} \right) \quad \text{otherwise}
%     \end{cases}
% \end{equation}
% % $$ a_n =  y_n p_{n-1} + a_{n-1} \left( 1 - p_{n-1} \right)$$\par

% % with $a_0 = \overrightarrow{0}$ and 
% $p_n$ is a monotonically decreasing function of the confidence scores defined as follows:
% \begin{equation}
%     \label{eq:1}
%     p_n = \prod_{i=1}^{n}h_{i} = h_{n} p_{n-1}
% \end{equation}

The model output is built inductively by using a monotonically decreasing function of the model confidence to interpolate between the current step's answer and the result of the same operation from the previous step.
We then train the model to reduce the classification loss of the final output with a regularizer that induces a bias towards reduced computation. 
Unlike the regularizer used by~\citet{eyzaguirre2020differentiable}, we use the following:
\begin{equation}
    \label{eq:loss}
    % \hat{L} (\vx, \vy) = L (\vx, \vy) + \tau \lVert h_i \rVert_{L_{1}}
    \hat{L} (x, y) = L (x, y) + \tau \sum_{i=1}^{n}h_i
\end{equation}
where $\tau$ is a hyper-parameter used to moderate the trade-off between complexity and error.
We find empirically that our changes help convergence and further binarize the halting probabilities.

Notably, the formulation is end-to-end differentiable.
This allows to fine-tune the weights of the underlying backbone (\textit{i.e.} the Transformer and embedding layers) using a joint optimization with the process that trains the intermediate classifiers.

\vspace{-0.1cm}
\subsection{Dynamic Computation at Inference}
\vspace{-0.1cm}

By construction, the DACT algorithm allows us to calculate upper and lower bounds of each of the output classes after any computation step (\textit{i.e.} Transformer block).
At inference, execution halts once the predicted probabilities for the topmost class are shown to remain higher than that of the \textit{runner-up} class (and by extension, of any other class).
That is, the model stops executing additional blocks once it finds that doing so will not change the class with maximum probability in the output because the difference between the top class and the rest is insurmountable.
Therefore, the \textit{halting condition} remains the same as the original DACT formulation~\citep{eyzaguirre2020differentiable}. More details of the method can be find at Appendix \ref{appedix:method}.
% Mathematically, we prove that the difference is insurmountable by comparing the lower bound for the probability of the top class predicted after $n$ blocks ($\Pr(c^*, n)$) with the upper bound of the probability of the runner-up class ($\Pr(c^{ru}, n)$).
% Therefore, the \textit{halting condition} remains the same as in the original DACT formulation~\citep{eyzaguirre2020differentiable}:
% \begin{equation}
%     \label{eq:9}
%     \Pr(c^*, n)(1-p_n)^d \geq \Pr(c^{ru}, n) + p_n d
% \end{equation}
% where $\Pr(c^*, n)$ and $\Pr(c^{ru}, n)$ are the probability of predicted after $n$ blocks for the top and runner up classes respectively.

\vspace{-0.1cm}
\subsection{Training}
\vspace{-0.1cm}

The training of the module follows a two step process. 
First, the underlying Transformer model must be tuned to the objective task.
This ensures a good starting point onto which the DACT module can then be adapted to and speeds up convergence. 
This is followed by a second fine-tuning phase where the complete model is jointly trained for the task. 
% This contrast with existing methods for dynamic Transformers, where first only pre-training the backbone, and later freezing the backbone and just modifying the weights of the classifiers.
This differs slightly from existing dynamic Transformer methods, which first pre-train the backbone and then freeze it to modify only the classifier weights.

% This latter training phase, not only makes it possible to use the module to work adaptively, but also modifies the Transformer layers making them more suited to work together, while also generating new representations in the Transformer which are useful for the new lower computation scenario.

\vspace{-0.1cm}
\section{Results}

\begin{figure*}[t]
    \begin{minipage}{.5\linewidth}
	\centering
	\includegraphics[width=0.49\columnwidth]{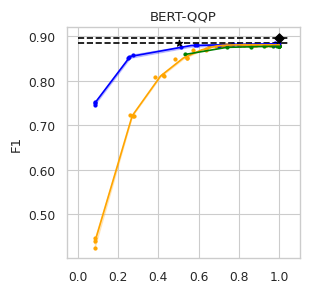}
	\includegraphics[width=0.49\columnwidth]{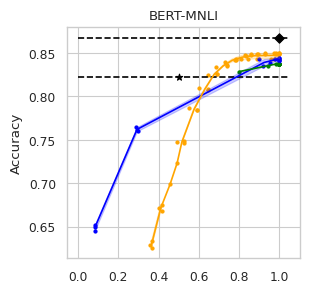}
	\includegraphics[width=0.49\columnwidth]{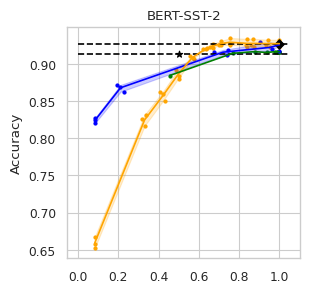}
	\includegraphics[width=0.49\columnwidth]{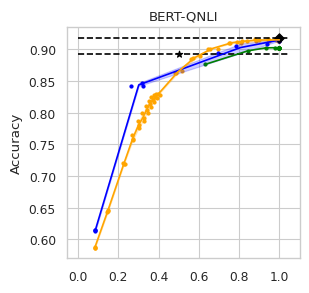}
	\includegraphics[width=0.49\columnwidth]{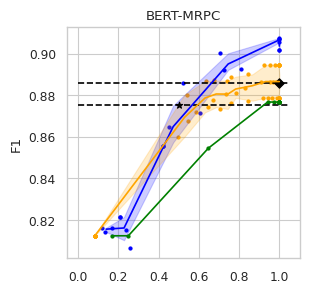}
	\includegraphics[width=0.49\columnwidth]{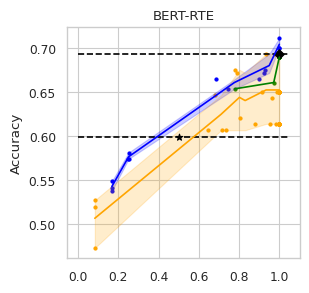}
    \end{minipage}
    \begin{minipage}{.5\linewidth}
	\centering
	\includegraphics[width=0.49\columnwidth]{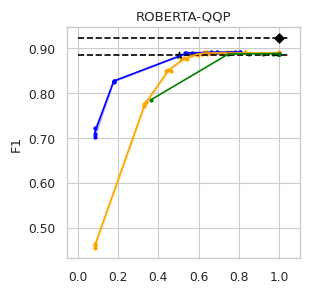}
	\includegraphics[width=0.49\columnwidth]{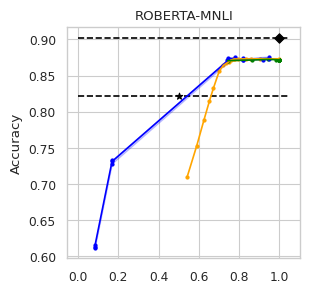}
	\includegraphics[width=0.49\columnwidth]{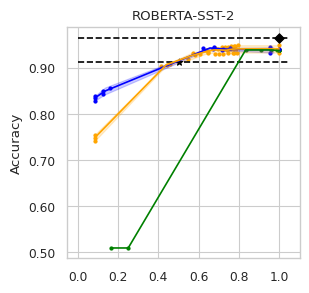}
	\includegraphics[width=0.49\columnwidth]{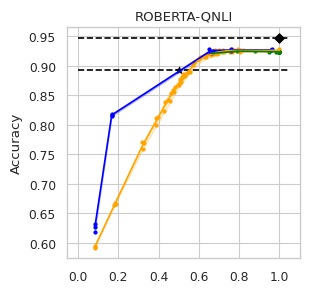}
	\includegraphics[width=0.49\columnwidth]{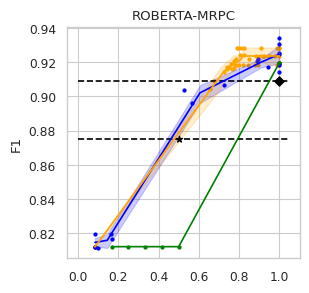}
	\includegraphics[width=0.49\columnwidth]{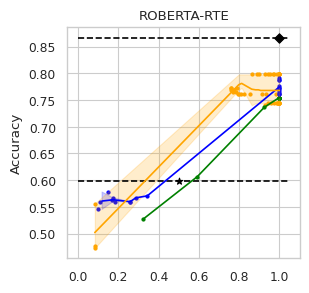}
    \end{minipage}
    % \vspace{-0.25cm}
	\caption[DACT-BERT diagram]{Performance vs efficiency trade-offs for BERT-base and RoBERTa-base models using {\color{dact-color}DACT-BERT (blue)},  {\color{deebert-color}DeeBERT (orange)} and {\color{pabee-color}PaBEE (green)}.
    DACT-BERT and DeeBERT experiments were repeated three times for each hyper-parameter. Individual runs are shown with colored dots, and the average along with its confidence interval is shown using a band.
	In all figures the \textbf{x-axis} shows computation measured as the fraction of the layers used by the respective static backbone (shown as a black diamond).
	\textbf{DistilBERT's} relative perfomance is shown at the 50\% computation mark using a black star.
	}
	\label{fig:result-figures}
    \vspace{-0.25cm}   
\end{figure*}

% \vladimir{tratemos de que la ultima pagina completa sea para resultados y conclusion}

% \begin{figure}
%     \begin{minipage}{.5\linewidth}
%     \centering
%     \subfloat[MRPC]{\label{main:a}\includegraphics[scale=.5]{images/entropy_mrpc.png}}
%     \end{minipage}
%     \begin{minipage}{.5\linewidth}
%     \centering
%     \subfloat[QNLI]{\label{main:b}\includegraphics[scale=.5]{images/entropy_qnli.png}}
%     \end{minipage}
%     \vspace{-1.4em}
    
%     \centering
%     \subfloat[RTE]{\label{main:c}\includegraphics[scale=.5]{images/entropy_rte.png}}
    
%     \definecolor{dact-color}{RGB}{ 55, 88, 218}
%     \definecolor{deebert-color}{RGB}{198, 76,84}
%     \caption{Attention entropy distribution per layer in the backbone for {\color{dact-color}DACT-BERT (blue)} and {\color{deebert-color}DeeBERT (red)} for three different GLUE tasks. Each point represent the entropy for one attention head in each layer and the line shows the mean entropy for all the attentions in a given layer.}
%     \label{fig:main}
% \end{figure}

\vspace{-0.1cm}
\subsection{Experimental Setup}
\vspace{-0.1cm}

We tested our method using both BERT and RoBERTa backbones, evaluating both models on six different tasks from the GLUE benchmark \citep{wang-etal-2018-glue}. We use DeeBERT \citep{xin2020deebert} and PABEE \citep{zhou2020bertlosespatience} as our dynamic baselines, using the same backbones for a fair comparison, and the respective non-adaptive backbones along with DistilBERT \citep{sanh2020distilbert} as static baselines. More details are presented at the Appendix \ref{appendix:implementation-details}.
% Our model was developed using PyTorch \citep{pytorch} on top of the implementation released by \citet{xin2020deebert} as well as the HuggingFace Transformers library \citep{Wolf2019HuggingFacesTS}\footnote{Code will be released upon publication.}.
% Because the focus of this paper was to introduce a new more-interpretable architecture and not achieve state of the art results we use the default parameters and architectures from the Transformers library \citet{Wolf2019HuggingFacesTS}

\vspace{-0.1cm}
\subsection{Computational Efficiency}
\vspace{-0.1cm}

To compare the trade-off that exists between computation efficiency and the performances obtained with it, we computed efficiency-performance diagrams for the validation set. Efficiency was measured as the proportion of Transformer layers used compared to the total number of layers in their static counterparts.
The specific metrics for performance are those suggested in the GLUE paper \citep{wang-etal-2018-glue} for each task.
% TO FIGURE
% Here the average amount of computation is shown in the horizontal axis, which is computed by the percentage of total transformer blocks executed. And the performance, which varies depending to the task, achieved in the vertical axis. The closer the curve is to the top left, the better.

In our experiments we fine-tune the backbone model for the GLUE tasks using the default values of the hyper-parameters. 
For the second stage we vary the value of $\tau$ in Equation \eqref{eq:loss} to compute our computation-performance diagram curves, selecting from a set of fixed values for all the experiments: $\tau \in \{5 \cdot 10^{-5}, 5 \cdot 10^{-4}, 5 \cdot 10^{-3}, 5 \cdot 10^{-2}, 5 \cdot 10^{-1}\}$. By modifying this hyperparameter in DACT we can manage the amount of computation the model will perform and record the performance it managed to achieve at this level. 

Similarly, using DeeBERT to create the computation-performance diagrams the entropy threshold was varied continuously in increments of $0.05$. For PaBEE we ﬂuctuate the patience value between 1 and 12, effectively trying out the full range. The results for the unmodiﬁed static backbones are also included as a reference, as are the results obtained by the half-depth DistilBERT pre-trained model.

The area under the curve (AUC) in the Performance vs. Efﬁciency plot shown in Figure~\ref{fig:result-figures} shows our approach improves the trade-off between precision and computation.
As was to be expected, all models perform similarly when saving little computation as they replicate the results achieved by the non-adaptive BERT backbone that performs a similar number of steps.
On the other hand, when using limited amounts of computation our model outperforms the alternatives in almost every task, especially in tasks for with more training data available.
We attribute this advantage in trading off computation and performance to ﬁne-tuning of the backbone weights for reduced computation.
Intuitively, as we move away from the 12 step regime for which the underlying static model was trained, more modiﬁcation of the weights is required.
Recall that of all the Dynamic Transformer algorithms only DACT-BERT can modify the Transformer weights because of its full-differentiability.
% Finally, we note that adaptive methods often offer the same performance at less computational cost than static ones.

% The results of this evaluation can be seen in Figure~\ref{fig:result-figures}.
% % Both results for the original BERT and RoBERTa performance were included.
% As was to be expected, both DACT-BERT and DeeBERT models perform similarly when using close to no efficiency savings as they replicate the results achieved by the non-adaptive BERT backbone.
% On the other hand, when using limited amounts of computation our model outperforms DeeBERT in almost every task (with a single exception).
% Essentially, this shows that for reducing computation in pre-trained transformers beyond a few percent our model is the clear choice.
% We attribute this advantage in trading off computation and performance to the combined effect of fine-tuning the backbone weights for reduced computation, and using a more robust adaptive mechanism.
% % On the other hand, our model almost in every task (except BERT on MRPC) outperforms DeeBERT when using a limited amount of computation. DACT-BERT can be more efficient than DeeBERT without compromising it's performance as much as DeeBERT does. 

Importantly, because our model learns to regulate itself, it shows remarkable stability in the amount of calculation saved. As the same values of ponder penalties give rise to similar efficiency outputs. By contrast, DeeBERT proves to be highly sensitive to the chosen value for the entropy hyperparameter.
% , exhibiting important fluctuations in both computation and performance indicators for small changes in its value (see RTE).
% Our robustness advantage seems to come from learning the efficiency mechanism instead of relying on a somewhat arbitrary heuristic for its control.
The robustness of our model appears to come from learning the efficiency mechanism rather than relying on a somewhat arbitrary heuristic for its control.

\vspace{-0.1cm}
\section{Conclusions}
\vspace{-0.1cm}

% This work explored the value of using the DACT algorithm with pre-trained Transformer architectures.
% Our results show that the proposed model, DACT-BERT, outperforms the current dynamic Transformer architectures in several tasks when significantly reducing computation.
% Moreover, our approach is also fully differentiable, a theoretical property necessary for most model interpretability methods.
% We hope that both qualities benefit future work on efficiency and transparency.

This work explored the value of using the DACT algorithm with pre-trained Transformer architectures. 
This results in a fully differentiable model that explicitly learns how many Transformers blocks it needs to perform a specific task.
Our results show that our approach, DACT-BERT, outperforms the current dynamic Transformer architectures in several tasks when significantly reducing computation.

% Entries for the entire Anthology, followed by custom entries
\bibliography{emnlp2021}
\bibliographystyle{acl_natbib}

\newpage

\appendix
\label{sec:appendix}

\section{Method Description \label{appedix:method}}

As illustrated in Figure~\ref{diagram} and shown in Algorithm \ref{algo:adaptive_computation}, we define the discrete unit of computation to be a single BERT Transformer block, \ie our gating mechanism will trade precision for additional complexity in discrete units of full additional Transformer blocks.
In addition to the output vector $y_n$ with the predicted class probabilities, each $n$-th DACT module computes an accompanying scalar confidence score (or halting value) $h_n$.

Line 8 shows how the output vectors are combined using a function of the halting values ($p_n$) to obtain the final predicted probabilities.
The intermediate results, accumulated in auxiliary variables $a_n$, encode the models best guess after unrolling $n$ Transformer layers.
Then, during inference, the confidence scores are used to effectively reduce computation by avoiding running all the layers using the appropriate halting criterion.
Because no non-differentiable functions are used during training the algorithm is end to end differentiable.

\begin{algorithm}
	\caption{\textit{DACT}} \label{algo:adaptive_computation}
	\begin{algorithmic}[1]
		\Require $M$ model with $N$ blocks
		\Require $is\_training \in \{ \True,\False \}$ 
        \State $p_n \leftarrow 1$
        \State $a_n \leftarrow \vec{0}$
		\For {step $n =1,2,\ldots N$}
		    \State \textcolor{blue}{\textit{\# Get output and confidence}}
            \State $y_n \leftarrow GetOutputModule(M, n)$        %\Comment{Run $n$th module}
            \State $h_n \leftarrow GetHaltModule(M, n)$          %\Comment{Get module confidence}

		    \State \textcolor{blue}{\textit{\# Combine with previous outputs}}
            \State $a_n \leftarrow (y_n * p_n) + (a_n * (1-p_n))$

            \State \textcolor{blue}{\textit{\# Update halting probability}}
            \State $p_n \leftarrow p_n * h_n$

            \State \textcolor{blue}{\textit{\# Stop computation during inference}}
		    \If{  not $is\_training$}
		        \If{  $AnsCantChange()$ }
		            \State \textbf{break loop}
                \EndIf
            \EndIf
		\EndFor
		\Ensure  Approximate final answer $a_n$
	\end{algorithmic}
\end{algorithm}

The inductive formulation of $a_n$ lends itself to calculating upper and lower bounds on the probabilities of the output classes.
If, during inference, it can be determined that the lowest possible value for the top-class $c^*$ in $a_n$ after running all $d$ remaining steps is still higher than the highest value for the runner-up class $c^{ru}$, then halting doesn't change the output:
\begin{equation}
    \label{eq:9}
    \Pr(c^*, n)(1-p_n)^d \geq \Pr(c^{ru}, n) + p_n d
\end{equation}

\section{Implementation Details \label{appendix:implementation-details}}

Our model was developed using PyTorch \citep{pytorch} on top of the implementations released by \citet{xin2020deebert} and \citet{zhou2020bertlosespatience}, as well as the HuggingFace Transformers library \citep{Wolf2019HuggingFacesTS}.
% Because the focus of this paper was to introduce a new more-interpretable architecture and not achieve state of the art results we use the default parameters and architectures from the Transformers library.
Because the focus of this paper was to introduce an alternative architecture of dynamic Transformers and not achieve state of the art results we use the default parameters and architectures from the Transformers library.

Both DeeBERT and DACT-BERT experiments were repeated three times to obtain the confidence intervals (95\% confidence) shown in Figure \ref{fig:result-figures}, each time using a different random initialization for the weights in the auxiliary classifiers \footnote{The random seeds were saved and will be published along with the code to facilitate replicating the results.}.
Results for FastBERT \citep{liu2020fastbert} are not reported since both DeeBERT and FastBERT use the same entropy-threshold halting criterion.
% , plus the addition of distillation in the latter.
% In fact, the main difference between them is the use of distillation in FastBERT, which can also be used with our proposed model, but this escapes the scope of our short paper.

Each experiment was run using a single 11GB NVIDIA graphics accelerator, which allows for training on the complete batch using 32-bit precision and without needing gradient accumulation.

\section{Compressibility}

\begin{figure}[ht]
    \vspace{-0.4cm}
    \centering
    \includegraphics[width=0.95\columnwidth]{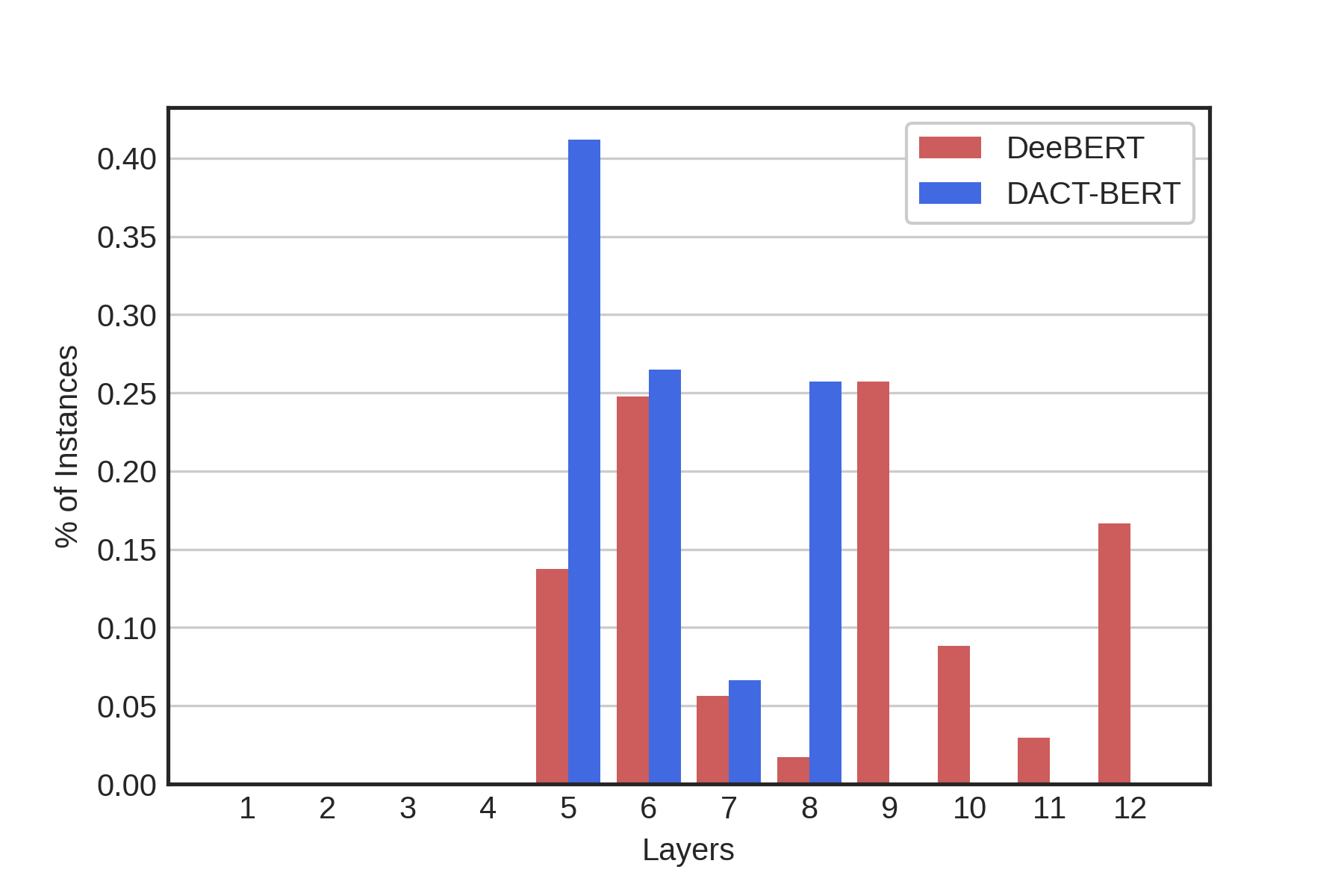}
    \vspace{-0.4cm}
    \caption{Frequency each Transformer block is used.}
    \label{fig:my_label}
    \vspace{-0.5cm}
\end{figure}

We find our model uses less layers compared to DeeBERT (see example at Fig. \ref{fig:my_label}), allowing us to prune the final layers.
We explain this difference by noting that the entropy will remain high throughout the whole model for the case of difficult questions as it will be uncertain about the answer.
On the other hand, any layer in DACT-BERT is capable of quitting computation if it believes future layers cannot answer with more certainty than its own (regardless of how certain the model actually is).

\end{document}